\documentclass{article}

\usepackage[accepted]{icml2018}

\usepackage[utf8]{inputenc} 
\usepackage[T1]{fontenc}    
\usepackage{hyperref}       
\usepackage{url}            
\usepackage{booktabs}       
\usepackage{amsfonts}       
\usepackage{amsmath}
\usepackage{graphicx}
\usepackage{listings}
\usepackage{subcaption}
\usepackage{microtype}      
\usepackage{xcolor}
\colorlet{lightyellow}{yellow!80}

\icmltitlerunning{Differentiable plasticity}

\begin{document}

\twocolumn[
\icmltitle{Differentiable plasticity: training plastic neural networks with backpropagation}

\begin{icmlauthorlist}
\icmlauthor{Thomas Miconi}{uber}
\icmlauthor{Jeff Clune}{uber}
\icmlauthor{Kenneth O. Stanley}{uber}
\end{icmlauthorlist}

\icmlaffiliation{uber}{Uber AI Labs}

\icmlcorrespondingauthor{Thomas Miconi}{tmiconi@uber.com}

\icmlkeywords{Machine Learning, ICML}
\vskip 0.3in
]

\printAffiliationsAndNotice{}  

\setlength{\belowcaptionskip}{-8pt}



\begin{abstract}
    How can we build agents that keep learning from experience, quickly and
    efficiently, after their initial training? Here we take inspiration from
    the main mechanism of learning in biological brains: synaptic plasticity,
    carefully tuned by evolution to produce efficient lifelong learning. We
    show that plasticity, just like connection weights, can be optimized by
    gradient descent in large (millions of parameters) recurrent networks with
    Hebbian plastic connections. First, recurrent plastic networks with more than two million
    parameters can be trained to memorize and reconstruct sets of novel,
    high-dimensional (1{,}000+ pixels) natural images not seen during training. Crucially,
    traditional non-plastic recurrent networks fail to solve this task. Furthermore, trained plastic networks can also solve generic meta-learning tasks such as the Omniglot task, with competitive results and little parameter overhead. Finally, in reinforcement learning settings, plastic networks outperform a non-plastic equivalent in a maze exploration task. We conclude that differentiable plasticity may provide a powerful novel approach to the learning-to-learn problem.

\end{abstract}

\section{Introduction: the problem of ``learning to learn''}

Many of the recent spectacular successes in
machine learning involve learning one complex task very well, through
extensive training over thousands or millions of training examples \cite{krizhevsky2012imagenet,mnih2015human,silver2016mastering}.
After learning is complete, the agent’s knowledge is fixed and
unchanging; if the agent is to be applied to a different task, it must be
re-trained (fully or partially), again requiring a very large number of
new training examples. By contrast, biological agents
exhibit a remarkable ability to learn quickly and efficiently from
ongoing experience: animals can learn
to navigate and remember the location of (and quickest way to) food sources,
discover and remember rewarding or aversive properties of novel objects and
situations, etc. -- often from a single exposure.

Endowing artificial agents with lifelong learning abilities is essential to
allowing them to master environments with changing or unpredictable features, or specific features that are unknowable at the time of training. For example,
supervised learning in deep neural networks can allow a neural network to
identify letters from a specific, fixed alphabet to which it was exposed during its training;
however, autonomous learning abilities would allow an agent to acquire knowledge of \emph{any}
alphabet, including alphabets that are unknown to the human designer at the time of training.

An additional benefit of autonomous learning abilities is that in many tasks (e.g.\ object recognition, maze
navigation, etc.), the bulk of fixed,
unchanging structure in the task  can be stored in the fixed knowledge of the agent, leaving
only the changing, contingent parameters of the specific situation to be
learned from experience. As a
result, learning the actual specific instance of the task at hand (that is,
the actual latent parameters that do vary across multiple instances of the
general task) can be extremely fast, requiring only few or even a
single experience with the environment. 


Several meta-learning methods have been proposed to train agents to learn autonomously (reviewed shortly). However, unlike in current approaches, in biological brains long-term learning is thought to occur \cite{martin2000synaptic,liu2012optogenetic}
primarily through \emph{synaptic plasticity} -- the strengthening and weakening of
connections between neurons as a result of neural activity, as carefully tuned
by evolution over millions of years to enable efficient learning during the
lifetime of each individual. 
While multiple forms of synaptic plasticity exist,
many of them build upon the general principle known as Hebb’s rule: if a neuron
repeatedly takes part in making another neuron fire, the connection between
them is strengthened (often roughly summarized as ``neurons that fire
together, wire together'') \cite{hebb1949organization}. 

Designing neural networks with plastic connections has long been explored with
evolutionary algorithms (see \citealt{soltoggio2017born} for a recent review), but has been so far relatively less studied in deep learning.
However, given the spectacular results of gradient descent in designing
traditional non-plastic neural
networks for complex tasks, it would be of great interest to expand
backpropagation training to networks with plastic connections -- optimizing through gradient descent not only the base weights, but also the amount of plasticity in each connection.



We previously
demonstrated the theoretical feasibility and analytical derivability of this
approach \cite{miconi2016backpropagation}. Here we show that this approach can train large
(millions of parameters) networks for non-trivial tasks. To demonstrate our approach, we apply it to three different types of tasks: complex pattern memorization (including natural images), one-shot classification (on the Omniglot dataset), and reinforcement learning (in a maze exploration problem). We show that plastic networks provide competitive results on Omniglot, improve performance in maze exploration, and outperform advanced non-plastic recurrent networks (LSTMs) by orders of magnitude in complex pattern memorization.
This result is interesting not only for opening up a new avenue of investigation in gradient-based neural network training, but also for showing that meta-properties of neural structures normally attributed to evolution or a priori design are in fact amenable to gradient descent, hinting at a whole class of 
heretofore unimagined meta-learning algorithms.

\section{Differentiable plasticity}

To train plastic networks with backpropagation, a plasticity rule must be specified. Many
formulations are possible. Here we choose a flexible formulation that 
keeps separate plastic and non-plastic (baseline) components for each
connection, while allowing multiple Hebbian rules to be easily implemented within the framework.

A connection between any two neurons $i$ and $j$ has both a fixed component and a plastic component.
The fixed part is just a traditional connection weight $w_{i,j}$. The plastic part
is stored in a \emph{Hebbian trace} $\mathrm{Hebb}_{i,j}$, which varies during
a lifetime
according to ongoing inputs and outputs (note that we use ``lifetime'' and ``episode'' interchangeably). In the simplest case studied here, the
Hebbian trace is simply a running average of the product of pre- and
post-synaptic activity.
The relative importance of plastic and fixed components in the connection is structurally determined by the plasticity coefficient $\alpha_{i,j}$, which multiplies the Hebbian trace to form the full plastic component of the connection.
Thus, at any time, the total, effective weight of the connection  between neurons $i$ and $j$ is the sum
of the baseline (fixed) weight $w_{i,j}$, plus the Hebbian trace $\mathrm{Hebb}_{i,j}$ multiplied by the
plasticity coefficient $\alpha_{i,j}$. The precise network equations for the output $x_j(t)$ of neuron $j$ are:
\begin{multline}
    x_j(t) = \sigma \big\{\sum_{i \in inputs}  [ w_{i,j}  x_i(t-1) \\ + \alpha_{i,j} \mathrm{Hebb}_{i,j}(t)  x_i(t-1) ] \big\},
\end{multline}
\begin{equation}
    \mathrm{Hebb}_{i,j}(t+1) = \eta  x_i(t-1) x_j(t) + (1 - \eta) \mathrm{Hebb}_{i,j}(t).  \label{eq:hebb}        
\end{equation}

Here $\sigma$ is a nonlinear function (we use $\tanh$ throughout this paper), and ``inputs'' denotes the set of all neurons providing input to neuron $j$. 

In this way, depending on the values of $w_{i,j}$ and $\alpha_{i,j}$, a connection can
be fully fixed (if $\alpha=0$), or fully plastic with no fixed component (if
$w=0$), or have both a fixed and a plastic component. 

The Hebbian trace $\mathrm{Hebb}_{i,j}$ is initialized to zero at the beginning of each
lifetime/episode: it is purely a lifetime quantity. The parameters $w_{i,j}$
and $\alpha_{i,j}$, on the other hand, are the
structural parameters of the network that are conserved across lifetimes, and
optimized by gradient descent between lifetimes (descending the gradient of the error computed during episodes), to maximize expected performance over a lifetime/episode.
Note that $\eta$, the ``learning rate'' of plasticity, is also an optimized parameter of the network. For simplicity, in this paper, all connections share the same value of $\eta$, which is thus a single, learned scalar parameter for the entire network. 

In Equation \ref{eq:hebb}, $\eta$ appears as a weight decay term, to prevent runaway positive feedback on Hebbian traces.
However, because of this weight decay, Hebbian traces (and thus memories) decay to zero in the absence of input. Fortunately, other, more complex Hebbian rules can maintain stable weight values indefinitely in the absence of stimulation, thus allowing stable long-term memories, while still preventing runaway divergences. One well-known example is Oja's rule \cite{oja2008oja}. To incorporate Oja's rule in our framework, we can simply replace Equation \ref{eq:hebb} above with the following (note the absence of a decay term):
\begin{multline}
    \mathrm{Hebb}_{i,j}(t+1) = \mathrm{Hebb}_{i,j}(t) \\ + \eta x_j(t) (x_i(t-1) - x_j(t) \mathrm{Hebb}_{i,j}(t)). \label{eq:oja}
\end{multline}

This method can train networks to form memories with arbitrary duration. To illustrate the flexibility of our approach, we demonstrate both rules in the experiments reported below. 

All experiments reported here use the PyTorch package to compute gradients. An important aspect of differentiable plasticity is its extreme ease of implementation, requiring only a few additional lines of code on top of a standard PyTorch network implementation. See Supplementary Materials for code snippets. The code for all experiments described in this paper is available at \url{https://github.com/uber-common/differentiable-plasticity}


\section{Related work}

Designing agents that can learn from ongoing experience is the basic problem of
meta-learning, or ``learning-to-learn'' \cite{thrun98learning}. 
Several methods already exist to address this problem. 
A straightforward approach is simply to train standard recurrent neural networks (RNNs) to adequately incorporate past experience in their future responses within each episode. Since RNNs are universal Turing
machines, they can \emph{in principle} learn any computable function of their
inputs. With a proper training schedule (e.g.\ augmenting inputs at time $t$ with the output and error at time $t-1$), recurrent networks can learn to automatically integrate novel information during an episode \cite{hochreiter2001learning,wang2016learning,duan2016rl2}.

To augment learning abilities, recurrent networks can be endowed with external content-addressable memory banks, as in Memory Networks and Neural
Turing Machines \cite{graves14neural,sukhbaatar15end,santoro16one}. The memory bank can be read from and written to by an attentional mechanism within the controller network,  enabling fast memorization of ongoing experience.

A different approach consists in augmenting each weight with a plastic component that automatically grows and decays as a (usually Hebbian) function of inputs and outputs. In our framework, this method is essentially equivalent to a plastic network in which all connections have the same, \emph{non-trainable} plasticity (i.e. identical and non-learnable $\alpha$, $\eta$, etc.): only the  non-plastic weights of the network are trained. \citet{schmidhuber1993reducing} has pointed out that such homogenous-plasticity networks can in principle learn to produce any desired trajectory. The recent ``fast-weights'' approach \cite{ba2016fast}, published concurrently with initial reports on differentiable plasticity \cite{miconi2016backpropagation}, augments recurrent networks with fast-changing Hebbian weights (all connections having the same, non-trainable plasticity) and computes activations iteratively at each time step (initializing each such loop with outputs of the slow-weighted, non-plastic network). The overall effect is to emphasize recently-encountered patterns, allowing the network to ``attend to the recent past'' \cite{ba2016fast}. 

Alternatively, one can optimize the learning rule itself, rather than the plasticity of connections. \citet{bengio1991learning} use a parametrized learning rule and optimize over these parameters (by meta-learning over multiple tasks), while the structure of the network (including the plasticity of each connection) is fixed a priori by the experimenter. Note that, similarly, our framework also includes limited rule optimization, because we optimize the rule parameter $\eta$.


 Perhaps the most general method is to have all weight updates be \emph{arbitrarily} computed on-the-fly by the network itself \cite{schmidhuber1993self}, or by a separate network \cite{schlag2017gated}, at each time step. This method is of course extremely flexible, but imposes a large learning burden on the networks.

Yet another approach is the MAML method \cite{finn2017model}, which performs gradient descent via backpropagation during the episode itself. In this case, the meta-learning consists in training the base network so that it can be ``fine-tuned'' easily and reliably by few steps (or even just one step) of additional gradient descent while performing the actual task.

For classification problems, one may instead train an embedding to reliably discriminate between ``different'' classes, as defined by the task. Then, during each episode, classification is reduced to a comparison between the embedding of the test and example instances. This approach is exemplified both by Matching Networks \cite{vinyals2016matching} (where the prediction of the test instance label is the sum of example instance labels, weighted by the cosine similarity of their embeddings to that of the test instance) and by Prototypical Networks \cite{snell2017prototypical} (in which the embedded examples are averaged to produce prototypical vectors for each class, with which the test instance is matched by nearest-neighbor classification).

One advantage of using trainable synaptic plasticity as a substrate for
meta-learning is the great potential flexibility of this approach. 
For example, Memory
Networks enforce a specific memory storage model in which memories must be
embedded in fixed-size vectors and retrieved through some attentional
mechanism. In contrast, trainable synaptic plasticity 
may translate into very different forms of memory, the exact implementation of which
can be determined by (trainable) network structure. Fixed-weight recurrent
networks, meanwhile, require neurons to be used for both storage and computation. While recurrent neural networks are universal Turing machines,
and can therefore perform any computable task \emph{in theory}, allowing the connections themselves to store information may reduce the computational burden on neurons.
Non-trainable plasticity networks (including fast-weights networks) can exploit network connectivity for storage of short-term information, but their uniform, non-trainable plasticity imposes a stereotypical behavior on these memories (``attending to the recent past'' \cite{ba2016fast}). By contrast, in natural brains, the amount and rate of plasticity in each connection are actively molded by 
the long-term meta-learning mechanism (namely, evolution) to perform specific computations and behaviors; furthermore, biological plasticity is not limited to fast weight changes, but can sustain memories over years or decades. Differentiable plasticity, by making the plasticity of each connection trainable, may thus provide a greater repertoire of behaviors and computations than non-trainable plasticity methods.

\section{Experiments and Results}

The experiments in this section are designed to show both that
differentiable plasticity actually works within a meta-learning framework, and that in some cases it provides a definitive
advantage over alternative options.  

\subsection{Pattern memorization: Binary patterns}

To demonstrate the differentiable plasticity approach, we first apply it to the task
of quickly memorizing sets of arbitrary high-dimensional patterns (including novel patterns never seen
during training), and reconstructing these patterns when exposed to partial,
degraded versions of them.  Networks that can perform this task are known
as content-addressable memories, or auto-associative networks \cite{dayan2001theoretical}.
This task is a useful test because it is known that hand-designed recurrent
networks with (usually homogenous) Hebbian plastic connections can successfully
solve it for binary patterns \cite{hopfield1982neural}. Thus, if differentiable 
plasticity is to be of any help, it should be able to automatically solve this task -- that
is, to automatically design networks that can perform the task just like
existing hand-designed networks can. 

\begin{figure}
    \centering
    \includegraphics[scale=.4]{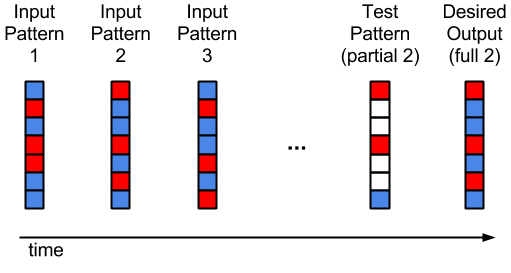}\hfill
    
    \includegraphics[scale=.66]{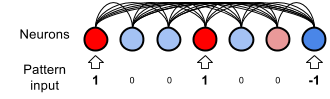}
    \caption{Binary pattern recognition task. Top: conceptual sequence. Bottom: depiction of the architecture.} \label{fig:schema}
\end{figure}

Figure \ref{fig:schema} (top) depicts an episode in this task. The network is shown a
set of 5 binary patterns in succession. Each binary pattern is composed of 1{,}000
elements, each of which is either 1 (dark red) or -1 (dark blue).
Each pattern is shown for 10 time steps, with 3 time steps of zero
input between presentations, and the whole sequence of patterns is presented 3
times in random order (few-shot learning). Then, one of the presented patterns is
chosen at random and degraded, by setting half of its bits to zero (i.e.\ no
input - white). This degraded pattern is then fed as an input to the network. The task
of the network is to reproduce the correct full pattern in its outputs, drawing on its
memory to complete the missing bits of the degraded pattern (pale blue and red in the bottom panel).

The architecture (Figure \ref{fig:schema}, bottom) is a fully recurrent neural network with one neuron per
pattern element, plus one fixed-output (``bias'') neuron, for a total of 1{,}001
neurons. Input patterns are fed by clamping the value of each neuron to the
value of the corresponding element in the pattern, if this value is not zero
(i.e. 1 or -1); for zero-valued inputs in
degraded patterns, the corresponding neurons do not receive pattern input, and
get their inputs solely from lateral connections, from which they must
reconstruct the correct, expected output values. Outputs are read directly from
the activation of the neurons.  The network’s performance is evaluated only on
the final time step, by computing the loss as the summed squared error between the final
network output and the correct expected pattern (that is, the non-degraded
version of the degraded input pattern). The gradient of this
error over the $w_{i,j}$ and $\alpha_{i,j}$ coefficients is then computed by
backpropagation, and these coefficients are optimized through an Adam solver \cite{kingma2015adam}
with learning rate 0.001.  For this experiment, we use the simple decaying Hebbian formula for updating Hebbian traces (equation \ref{eq:hebb}). Note that the network has two trainable parameters
($w$ and $\alpha$) for each connection, summing up to $1{,}001 \times 1{,}001
\times 2 = 2{,}004{,}002$ trainable parameters.

\begin{figure}[t]
\centering
    \includegraphics[scale=.5]{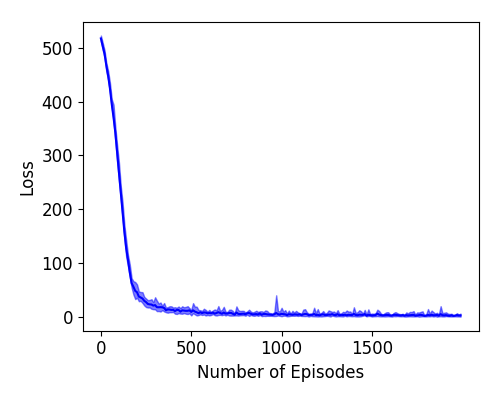}
\caption{Learning curve for 1{,}000-bit pattern memorization (10 runs shown:
        narrow shaded area indicates minimum and maximum loss, thick curve indicates
        mean loss).}    
        \label{fig:simple}
\end{figure}

Figure \ref{fig:simple} shows the result of 10 runs with different random seeds.
Error (defined as the proportion of bits that have the wrong sign) converges to a low, residual value (<1\%) within about 200 episodes.

\subsection{The importance of being plastic: a comparison with non-plastic
recurrent networks}

\begin{figure}
\centering
    \includegraphics[scale=.5]{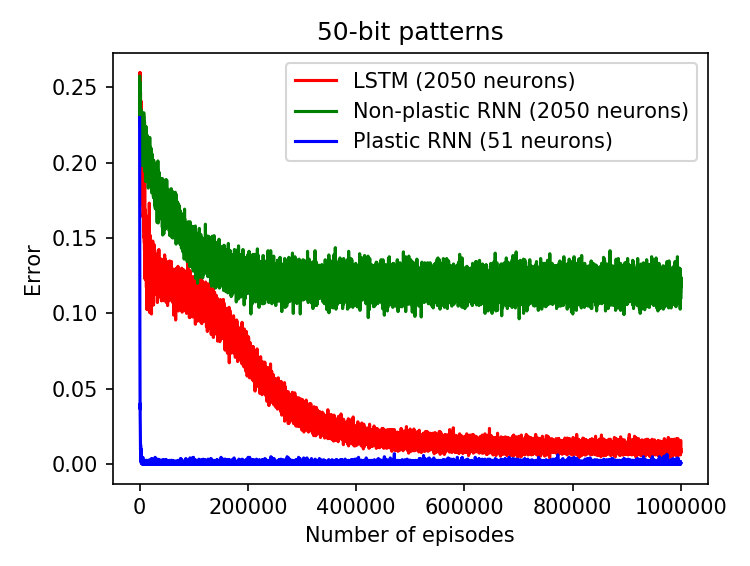}
        \caption{Learning curve for typical runs for 50-bit patterns, using a
        non-plastic RNN with 2,050 neurons (green curve), an LSTM with 2,050 neurons (red curve), and a differentiable plastic-weight
        network with the same parameters but only 51 neurons (blue curve).}
        \label{fig:comparison}
\end{figure}

In principle, this task (like any computable task) could be solved by a
non-plastic recurrent network, although the non-plastic networks will require additional neurons to store previously seen patterns.  However, despite much exploration, we were unable
to succeed in solving this task with a non-plastic RNN or LSTM \cite{hochreiter1997long}. We could
only succeed by reducing the pattern size to 50 bits (down from 1{,}000),
showing only 2 patterns per episode (rather than 5), and presenting them for only 3 time steps. The best results
required adding 2,000 extra neurons (for a total of 2,050 neurons). Training error
over episodes is shown in Figure \ref{fig:comparison}. For the non-plastic RNN, the error essentially flatlines
at a high level (green curve). The LSTM solves the task, imperfectly, after about 500{,}000 episodes (red curve). For comparison, the blue curve shows performance on the exact same
problem, architecture, and parameters, but restoring plastic connections.
The network solves the task very quickly, reaching mean error below .01 within ~2{,}000 episodes, which is 250 times faster than the LSTM. 

Thus, for this specific task, plastic recurrent networks seem considerably more powerful than LSTMs. Although this  task is known to be well-suited for plastic recurrent networks \cite{hopfield1982neural}, this result raises the question of which other domains might benefit from the differentiable plasticity approach over current LSTM models (or even by adding plasticity to LSTM models). 


\begin{figure*}[t]
\centering
    \begin{subfigure}{.57\textwidth}
        \includegraphics[width=.99\textwidth]{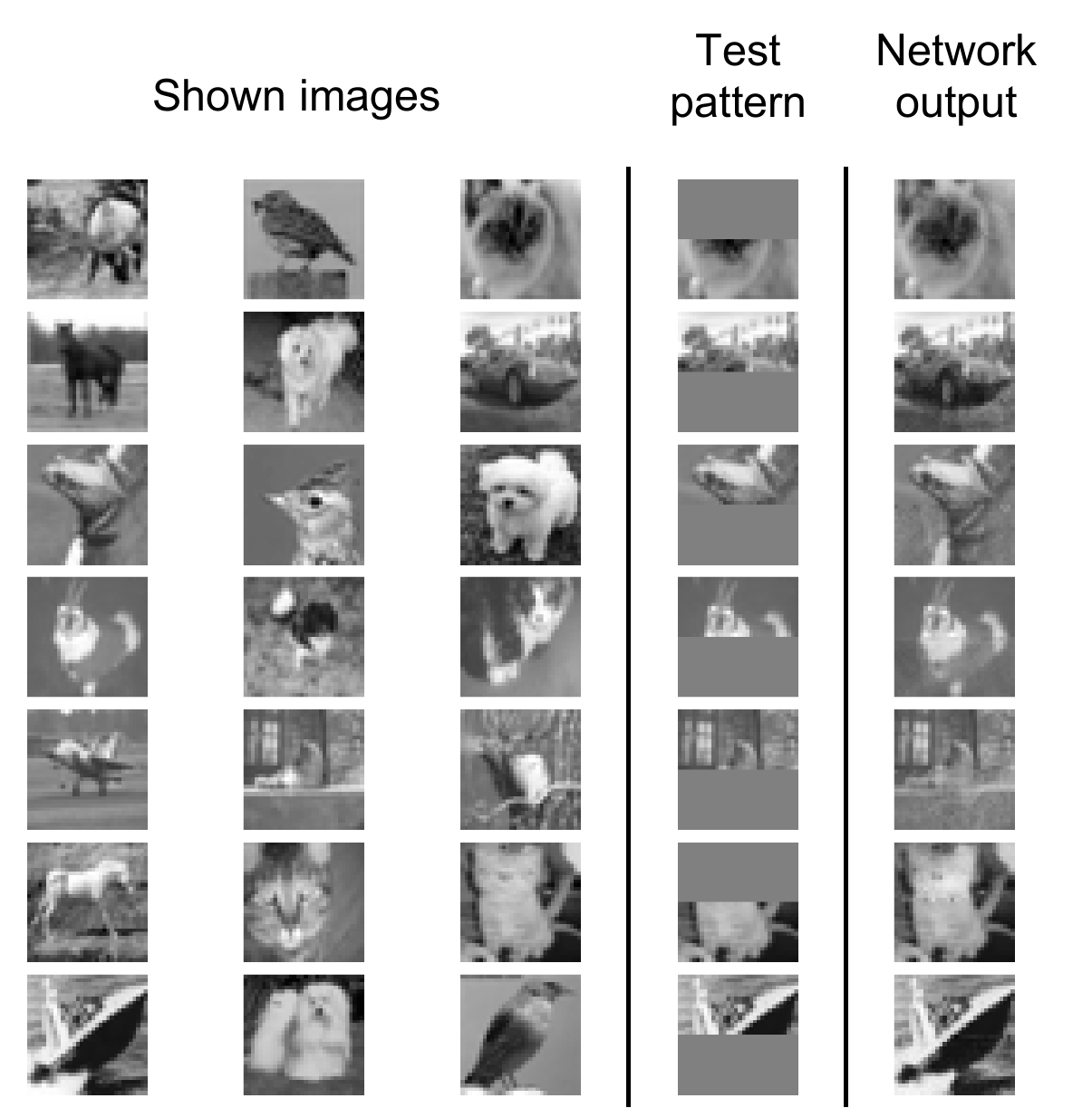}
        \caption{Typical image reconstruction results from a withheld test
    set (not seen during training). Each row is a full episode.}\label{fig:images}
    \end{subfigure}
    \begin{subfigure}{.42\textwidth}
         \includegraphics[width=1.0\textwidth]{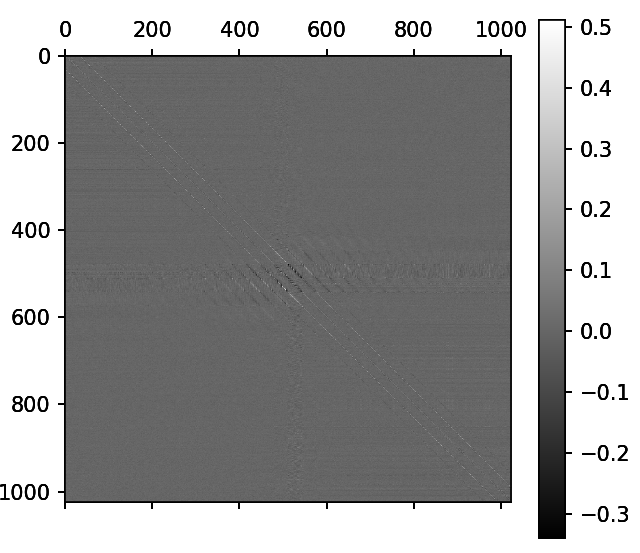} \\ \includegraphics[width=1.0\textwidth]{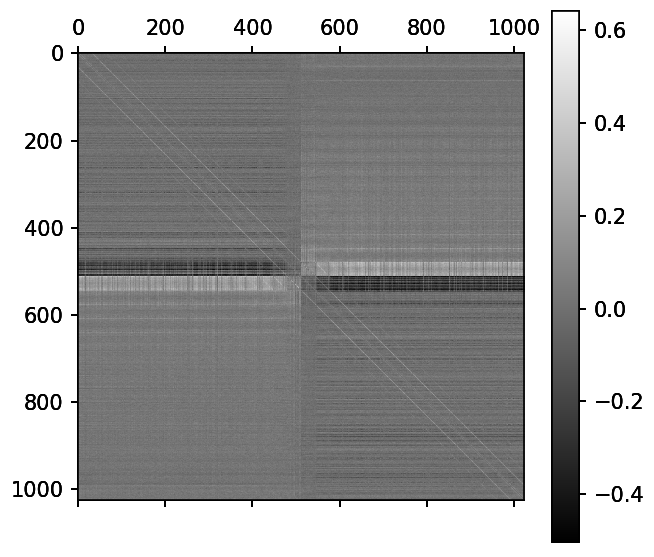} 
         \caption{Matrices of baseline weights
    $w_{i,j}$ (top) and plasticity coefficients $\alpha_{i,j}$ (bottom) after
    training. Each column describes the input to a single cell, and vertically
    adjacent entries describe inputs from horizontally adjacent pixels in the
    image. Notice the significant structure present in both matrices (best viewed electronically by zooming).}\label{fig:matrices}
    \end{subfigure}
\caption{Natural image memorization through differentiable plasticity.}
\end{figure*}

\begin{figure}
    \includegraphics[scale=.666]{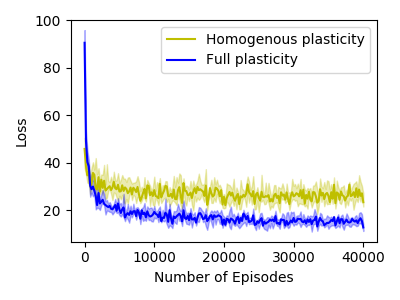}
        \caption{ Median and inter-quartile range over 10 runs for a plastic network with independent $\alpha$ (blue curve) or shared (homogenous) $\alpha$, equivalent to an optimal fast-weights network (yellow curve). Independent $\alpha$ coefficients for each connection improve performance, by allowing useful structure in the plasticity matrix.}
        \label{fig:yoked}
\end{figure}

\subsection{Pattern memorization: Natural images}

As a more challenging test, we applied our method to the problem of memorizing
natural images with graded pixel values, which contain much more information per
element. Images are from the CIFAR-10 database, which contains 60{,}000
images of size 32 by 32 pixels (i.e.\ 1{,}024 pixels in total), converted to
grayscale pixels between 0 and 1.0. The architecture is largely similar to the
one described above, with 1{,}025 neurons in total, leading to $2 \times 1{,}025
\times 1025 = 2{,}101{,}250$ parameters. Each episode included 3 pictures, shown
3 times (in random order each time) for 20 timesteps each time, with 3 time steps of zero input
between image presentations. To prevent a trivial solution consisting in simply
reconstructing each missing pixel as the average of its neighbors (which the
high autocorrelation of natural images might make viable), images are degraded by
zeroing out one full contiguous half of the image (either top or bottom half).

Figure \ref{fig:images} shows the behavior of the trained network on a
withheld test set of images not seen during training. The last column shows the
final output of the network, i.e.\ the reconstructed image.  The model has
successfully learned to perform the non-trivial task of memorizing and
reconstructing previously unseen natural images. 

Figure \ref{fig:matrices} shows the
final matrices of weights (top) and plasticity coefficients (bottom) produced by training. The
plasticity matrix (bottom) shows considerable structure, in contrast to the homogenous
plasticity of traditional Hopfield networks \cite{hopfield1982neural}. Some of
the structure (diagonal lines) is related to the high correlation of
neighboring pixels, while other aspects (alternating bands near the
midsection) result from the choice to use half-field zeroing in test images. We hypothesize that the wide alternating bands near the midsection support fast clearing of ongoing network activity when a test stimulus is presented (see Supplementary Materials, section \ref{section:plastic}).

While Figure \ref{fig:matrices} shows that the learned network has structure, it could be that this structure is merely an artifact of the learning process, with no inherent usefulness. To test this possibility, we compare the full plastic network against a similar architecture with shared plasticity coefficients -- that is, all 
connections share the same $\alpha$ coefficient. Thus plasticity is still trainable, but as a single parameter that is shared across all connections. Interestingly, because we use the simple decaying Hebbian formulation here, this shared-plasticity architecture has similarities with a fast-weights network \cite{ba2016fast}; however, unlike the fast-weights approach, differentiable plasticity allows us to learn both the importance of the fast weights ($\alpha$) and their learning rate ($\eta$) by gradient descent (also, we do not implement the iterative computation of neural activity at each time step used by fast-weights networks). The result of this comparison is shown in Figure \ref{fig:yoked}. The main outcome is that independent plasticity coefficients for each connection improve performance for this task. This comparison shows that the structure observed in Figure \ref{fig:images} is actually useful, and constitutes a novel architecture for memorization and reconstruction of natural images under these settings. 


\subsection{One-shot pattern classification: Omniglot task}


While fast pattern memorization is a complex task, it is important to assess whether differentiable plasticity can handle a wider range of tasks. To test this, we first apply our approach to the standard task for one-shot and few-shot learning, namely, the Omniglot task. 

The Omniglot dataset  \cite{lake2015human} is a collection of handwritten characters from various writing systems, including 20 instances each of 1,623 different handwritten characters, written by different subjects. This dataset is the basis for a standard one-shot and few-shot learning task, organized as follows: in each episode, we randomly select N character classes, and sample K instances from each class (here we use N=5 and K=1, i.e.\ five-way, one-shot learning). We show each of these instances, together with the class label (from 1 to N), to the model. Then, we sample a new, unlabelled instance from one of the N classes and show it to the model. Model performance is defined as the model's accuracy in classifying this unlabelled example.

Our base model uses the same base architecture as in previous work \cite{vinyals2016matching,finn2017model,snell2017prototypical,mishra2017simple} : 4 convolutional layers with $3 \times 3$ receptive fields and 64 channels. As in \citet{finn2017model}, all convolutions have a stride of 2 to reduce dimensionality between layers. The output of this network is a single vector of 64 features, which feeds into a N-way softmax. Concurrently, the label of the current character is also fed as a one-hot encoding to this softmax layer, guiding the correct output when a label is present. 

There are several ways to introduce plasticity in this architecture. Here, for simplicity, we chose to restrict plasticity solely to the weights from the final layer to the softmax layer, leaving the rest of the convolutional embedding non-plastic. Thus, across many training episodes, we expect the convolutional architecture to learn an adequate discriminant between arbitrary handwritten characters. Meanwhile, the plastic weights between the convolutional network and the softmax should learn to memorize associations between observed patterns and outputs, which are directly influenced by the labels when these are present. 
For this experiment, we use Oja's rule (Eq. \ref{eq:oja}) as it seemed to improve performance. Following common practice, we augment the dataset with rotations by multiples of 90$^{\circ}$. We divide the dataset into 1,523 classes for training and 100 classes (together with their augmentations) for testing. We train the networks with an Adam optimizer (learning rate $3 \times 10^{-5}$, multiplied by $2/3$ every 1M episodes) over 5{,}000{,}000 episodes. To evaluate final model performance, we train 10 models with different random seeds, then test each of those on 100 episodes using the (previously unseen) test classes. 

The overall accuracy (i.e. proportion of episodes with correct classification, aggregated over all test episodes of all runs) is 98.3\%, with a 95\% confidence interval of 0.80\%. The median accuracy across the 10 runs is 98.5\%, indicating consistency in learning. Table \ref{table:results} compares our results with those reported in recent papers. All these reports made use of the simple convolutional embedding described above, which was introduced by \citet{vinyals2016matching}. However, these approaches differ widely in terms of computational cost and number of parameters.
Our results are largely similar to those reported for the computationally intensive MAML method \cite{finn2017model} and the classification-specialized Matching Networks method \cite{vinyals2016matching}, and slightly below those reported for the SNAIL method \cite{mishra2017simple}, which trains a whole additional temporal-convolution network on top of the convolutional architecture described above, thus adding many more parameters. We conclude that simply adding a few plastic connections in the output of the network (for a total addition of $64 \times 5 = 320$ parameters
over $111{,}426$ parameters in the overall network) allows for competitive one-shot learning over arbitrary man-made visual symbols.

\begin{table}
\centering
\def\arraystretch{1.25}
\caption{Results for the 5-way, 1-shot omniglot tasks, including recent reported results and the new differentiable plasticity (DP) result ($\pm$ indicates 95\% CI). Note that these reports describe widely varying approaches and model sizes (see text).}
\label{table:results}
\begin{small}
\begin{tabular}{c|c} 
\toprule
 Memory Networks \cite{santoro16one} & 82.8\% \\
 Matching Networks \cite{vinyals2016matching} & 98.1\% \\
 ProtoNets  \cite{snell2017prototypical} & 97.4\% \\
 Memory Module \cite{kaiser2017learning} & 98.4 \% \\
 MAML \cite{finn2017model} &  98.7\% $\pm$ 0.4 \\
 SNAIL \cite{mishra2017simple} & 99.07\% $\pm$ 0.16 \\
 DP (Ours) & 98.3\% $\pm$ 0.80 \\
\bottomrule
\end{tabular}
\end{small}
\end{table}


\subsection{Reinforcement learning: Maze exploration task}

\begin{figure}[t]
\centering
\includegraphics[scale=0.2]{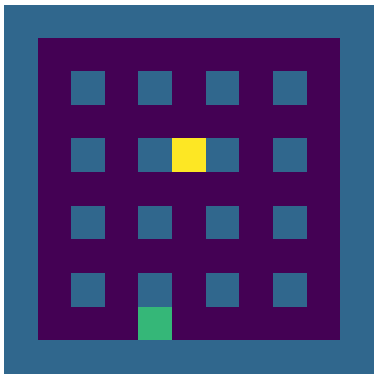}
\includegraphics[scale=0.48]{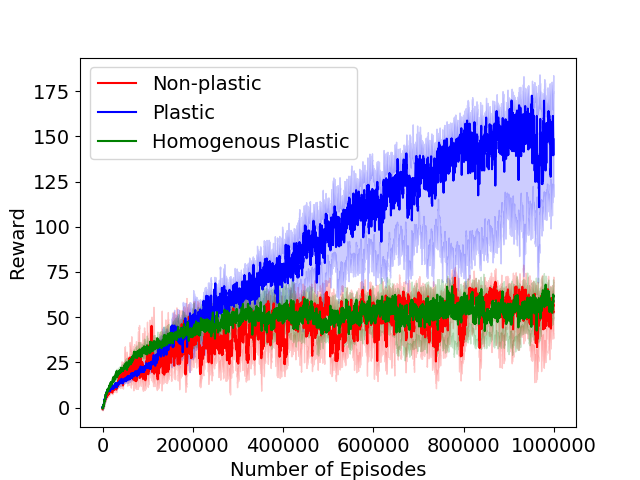}
\caption{Maze navigation reinforcement learning. Top: schematic representation of the maze, indicating the agent location (yellow) and the reward location (green, for illustration only: the reward is not visible to the agent). Bottom: Training curves for the maze exploration task: median and inter-quartile range of reward over 15 runs for each episode.} \label{fig:maze}
\end{figure}

Simple recurrent neural networks can be trained to perform reinforcement learning  tasks, in such a way that the network implements its own self-contained reinforcement learning algorithm during each episode \cite{wang2016learning,duan2016rl2}. Differentiable plasticity might improve the learning abilities of these networks. To test this possibility, we devise a simple maze exploration task. 

The maze is composed of $9 \times 9$ squares, surrounded by walls, in which every other square (in either direction) is occupied by a wall. Thus the maze contains 16 wall squares, arranged in a regular grid (Figure \ref{fig:maze}, top). The shape of the maze is fixed and unchanging over the whole task. At each episode, one non-wall square is randomly chosen as the reward location. When the agent hits this location, it receives a large reward (10.0) and is immediately transported to a \emph{random} location in the maze (we also provide a small negative reward of -0.1 every time the agent tries to walk into a wall). Each episode lasts 250 time steps, during which the agent must accumulate as much reward as possible. The reward location is fixed within an episode and randomized across episodes. Note that the reward is invisible to the agent, and thus the agent only knows it has hit the reward location by the activation of the reward input at the next step (and possibly by the teleportation, if it can detect it).

Inputs to the agent consist of a binary vector describing the $3 \times 3$ neighborhood centered on the agent (each element being set to 1 or 0 if the corresponding square is or is not a wall), together with the reward at the previous time step, following common practice \cite{wang2016learning}. The architecture is a simple recurrent network with 200 neurons, with a softmax layer on top of it to select between the 4 possible actions (up, right, left or down). As in \cite{wang2016learning}, we use the Advantage Actor-Critic algorithm(A2C, i.e. a single-threaded, non-parallel variant of the A3C policy gradient search algorithm \cite{mnih2016asynchronous}) 
to meta-train the network.

We run the experiments under three conditions: full differentiable plasticity, no plasticity at all, and homogenous plasticity in which all connections share the same (learnable) $\alpha$ parameter. Everything else is identical between all three conditions; we use Oja's rule throughout (Eq. \ref{eq:oja}).
For each condition, we perform 15 runs with different random seeds.  Note that the ``no-plasticity'' condition is conceptually identical to the method of \cite{wang2016learning}, that is, meta-training non-plastic RNNs with A2C.

As shown in Figure \ref{fig:maze} (bottom), differentiable plasticity strongly improves performance in this maze-learning task.
The curves suggest that simple RNNs get ``stuck'' on a sub-optimal strategy. Importantly, homogenous-plasticity networks also settle on a low performance plateau. This result suggests that in this domain (like in image completion) individually sculpting the plasticity of each connection is crucial in reaping the benefits of plasticity for this task.
\section{Discussion and conclusion}

The idea that an optimized form of plasticity can play a key role in learning is entirely natural, which makes it interesting that it is so rarely exploited in modern neural networks.
While supervised learning of fixed networks provide a powerful option for general learning, task-specific learning from experience may greatly improve performance on real-world problems.
Plasticity is literally the natural choice for this kind of learning. The results in this paper highlight that simple plastic models support efficient meta-learning.  Furthermore, they reveal for the first time that gradient descent itself (without the need for any evolutionary process) can optimize the plasticity of such a meta-learning system, raising the novel prospect that all the progress and power gained through recent years of deep learning research can be brought to bear on training and discovering novel plastic structures.

Our experiments show that this kind of meta-learning can vastly outperform alternative options on some tasks. Furthermore, in Omniglot, differentiable plasticity performed competitively, 
 despite its relative simplicity and compactness (in terms of added parameters), which suggests a unique profile of trade-offs for this novel approach. 

To gain a comprehensive understanding of the implications of this new tool will require a broad program of research. It is applicable to almost any meta-learning problem, and even outside meta-learning it could prove helpful for certain challenges.  For example, it might complement recurrent structures in arbitrary temporal or sequential domains.  Plastic LSTMs are conceivable future models.  Furthermore, the breadth of possible plasticity models is wide even on its own.
For example, neuromodulation (the control of moment-to-moment plasticity by network activity) has been shown to improve the performance of plastic neural networks designed by evolutionary algorithms \cite{soltoggio2008evolutionary}. Because our framework explicitly parametrizes plasticity, it is well-suited to exploring neuromodulatory approaches 
(for example by making $\eta$ or $\alpha$ depend in part on the activity of some neurons). Simple recurrent neural networks can be trained to perform reinforcement learning \cite{wang2016learning}, and the new results herein show that simple plasticity already improves learning performance on a simple such task; however, the highly elaborate system of neuromodulation implemented in animal brains 
\cite{frank2004carrot}
is unlikely to be accidental, and incorporating neuromodulation in the design of neural networks may be a key step towards flexible decision-making.
The prospects for this and other such ambitious enterprises are at least more conceivable now with the evidence that meta-learning plasticity is achievable through gradient descent.


\section*{Acknowledgements}

We thank David Ha for fruitful discussions and suggestions. We thank Juergen Schmidhuber and Yoshua Bengio for helpful references to previous work in plastic and self-modifying networks.

\bibliography{biblio}
\bibliographystyle{icml2018}

\pagebreak
\onecolumn
\begin{center}
\textbf{\large Supplementary Materials}
\end{center}
\setcounter{equation}{0}
\setcounter{figure}{0}
\setcounter{table}{0}
\setcounter{section}{0}
\makeatletter
\renewcommand{\theequation}{S\arabic{equation}}
\renewcommand{\thesection}{S\arabic{section}}
\renewcommand{\thefigure}{S\arabic{figure}}
\renewcommand{\bibnumfmt}[1]{[S#1]}
\renewcommand{\citenumfont}[1]{S#1}

\section{Code example for differentiable plasticity}

The code in Listing \ref{code:dp} shows how to implement differentiable plasticity with PyTorch. The code defines and meta-trains a simple recurrent neural network with plastic connections. The code specific to differentiable plasticity has been highlighted in yellow; without these highlighted passages, the code simply implements a standard RNN. Note that implementing differentiable plasticity requires less than four lines of additional code on top of a simple RNN implementation.

The full program from which this snippet is extracted is available at \url{https://github.com/uber-common/differentiable-plasticity/blob/master/simple/simplest.py}.

\begin{small} 
    \begin{lstlisting}[language=Python,escapechar=|,caption={Code snippet for meta-training a plastic RNN. Additional code to implement plasticity is highlighted in yellow.},label={code:dp}]
w = Variable(.01 * torch.randn(NBNEUR, NBNEUR), requires_grad=True) # Fixed weights
|\colorbox{lightyellow}{alpha = Variable(.01 * torch.randn(NBNEUR, NBNEUR), requires\_grad=True)}| # Plasticity coeffs. 
optimizer = torch.optim.Adam([w, alpha], lr=3e-4)

for numiter in range(1000): # Loop over episodes
    y = Variable(torch.zeros(1, NBNEUR)) # Initialize neuron activations
    |\colorbox{lightyellow}{hebb = Variable(torch.zeros(NBNEUR, NBNEUR))}| # Initialize Hebbian traces
    inputs, target = generateInputsAndTarget() # Generate inputs & target for this episode
    optimizer.zero_grad()
    # Run the episode:
    for numstep in range(NBSTEPS):
        yout = F.tanh( y.mm(w |\colorbox{lightyellow}{+ torch.mul(alpha, hebb))}| +
                Variable(inputs[numstep], requires_grad=False) )
        |\colorbox{lightyellow}{hebb = .99 * hebb + .01 * torch.ger(y[0], yout[0])}| # torch.ger = Outer product
        y = yout
    # Episode done, now compute loss, apply backpropagation
    loss = (y[0] - Variable(target, requires_grad=False)).pow(2).sum()
    loss.backward()
    optimizer.step()

\end{lstlisting}
\end{small}

\section{Details for the image reconstruction task}

Stimuli are natural images taken from the CIFAR10 dataset, of size $32 \times 32$ pixels. All images are normalized within the $[-1,1]$ range by subtracting the mean pixel value from each pixel and then dividing the resulting values by the maximum absolute pixel value.

The network is a fully-connected recurrent network of 1,025 neurons (one per image pixel, plus one ``bias'' neuron with output clamped to 1). Each of the non-bias neurons corresponds to one pixel. During each episode, three images are presented three times each in succession (within each of the three presentations, the three images are shown in random order). Each image is shown for 20 time steps, with 3 time steps of zero input between each image presentation. Then the test stimulus (one of the three images, with either the top or bottom half zeroed out and providing no input) is shown for 3 time steps. The error to be minimized by the network is the sum of squared difference between the pixel values in the (non-degraded) test image and the output of the corresponding neurons at the last time step.

Input is provided to the network by clamping the output of each input-receiving neuron to the value of the corresponding pixel. Zero pixels (i.e. the pixels in the blanked out portion of the test image) provide no input. At each time step, neurons not currently receiving image input update their activation according to Eq. 1, and the plastic component of each connection is updated according to Eq. 2. 

At the end of each episode, the loss was computed, and the gradient of this loss over the $\alpha_{i,j}$, $w_{i,j}$ and $\eta$ parameters was computed using PyTorch's \texttt{backward()} function. This gradient was then fed to an Adam optimizer (learning rate 1e-4) to update the parameters.

\section{Discussion of the plasticity structure in trained image-reconstructing networks}
\label{section:plastic}

The matrix of plasticity coefficients in the trained image-reconstructing network (Figure \ref{fig:matrices}, bottom) shows significant
structure. This pattern contrasts with traditional models of pattern completion in plastic recurrent
networks (such as Hopfield networks), which generally use homogenous plasticity across connections.
Here we suggest an explanation for this complex structure. 

In short, we propose that the network has learned a clever mechanism, not only to reconstruct the missing portion of the image, but also to clear away the remnant, unneeded activity from previous stimuli that might interfere with the reconstruction. This mechanism exploits the fact that partial images are always shown as top or bottom halves, and also the high autocorrelation of natural images.

At test time, if a neuron falls within the blank part of the degraded stimulus, it must determine its correct output from the activity of other neurons. For this purpose, it should take information from neurons from the other half, because they will receive the informative portion of the stimulus. This requirement is reflected in the homogenous domains of small, but significantly positive coefficients in the top-right and bottom-left quadrants of the matrix: neurons receive small positive plastic connections from the other half. Because these connections are nearly homogeneous, this portion of the network operates exactly like a standard Hopfield network (each plastic connection develops a weight that is proportional to the correlation between the two pixels across learned images,  providing a total input to each pixel that is roughly the regression of that pixel's value over the values of the source pixels).

The thin diagonal stripes of high, positive plasticity simply
reflect the high correlation between neighboring pixels that occurs in natural images: because neighboring neurons are highly correlated, it is useful for each pixel to receive information from its immediate neighbors.

The purpose of the large alternating bands near the mid-section is less 
obvious. The dark horizontal bands near the midsection mean that every pixel in the image receives
connections with a large, \emph{negative} plasticity coefficient from the farthest portion of its own half (i.e. the
bottom of the top half if the pixel is in the top half, or the top few of the bottom half if the pixel is in
the bottom half). Similarly, the pale bands show that each pixel receives connections with \emph{positive}
plasticity coefficients from the closest portion of the other half. What could be the purpose of this
arrangement?

Due to Hebbian plasticity, plastic connections store the correlation between the two pixels (positive or
negative), and thus at test time a highly plastic incoming connection will instruct the receiving neuron
to fire according to their observed historical correlation with the source neuron - ``how much
should I fire, given your current firing rate and our observed historical correlation during the episode?''

When the plasticity coefficient has negative sign, the opposite occurs: the source neuron not only
ignores, but acts contrarily to its expected firing based on current activity of the source neuron -- ``tell
me how much I should fire based on your current firing and our past correlation, and I'll do the
opposite!''

Therefore, we propose the following hypothesis: To adequately complete the partial pattern, the network must not
only fill the blank portion with adequate data -- it must also erase whatever information was already
there as a remnant of previous activity. That seems to be the purpose of the negative plasticity
coefficient from the extremum of the same half: at test time, if a neuron is within the half receiving the
informative portion of the stimulus, then this connection has no effect (because neural output is
clamped to the incoming stimulus). However, if they are in the blank half of the test stimulus (the one
that must be reconstructed), this negative-plastic connection will instruct the neurons to alter their
firing in opposition to what is predicted from the current value of their same-half brethren, thus actively
``forgetting'' whatever remnant activation is still echoing in this half of the network from previous
stimuli. Note that on its own, this ``cleaning'' mechanism might interfere with correct reconstruction,
e.g. when neurons start to settle on the correct reconstruction, this forced decorrelation might impair it. 
Yet that outcome is prevented by the counter-balancing positive-plasticity connections from the closest portion of
the other half. At test time, these opposite-half pixels will always contain correct information (because they receive the
informative part of the test pattern), while generally having very similar values to the nearby same-half pixels
sending out negative-plasticity connections (because the corresponding pixels are very close, falling
immediately on either side of the midsection). As a result, the ``cleaning'' effect will be strong when
distal same-half pixels differ from proximal other-half pixels, but will cancel out if they have similar
(correct) activity. This mechanism allows the network to actively discard junk activity while preserving accurate
reconstructions generated by the low-level, homogenous plastic connections from the entire opposite
half described above.

\end{document}